\documentclass{article}
\usepackage{kr}

\usepackage{times}
\usepackage{soul}
\usepackage{url}
\usepackage[hidelinks]{hyperref}
\usepackage[utf8]{inputenc}
\usepackage[small]{caption}
\usepackage{graphicx}
\usepackage{amsmath}
\usepackage{amsthm}
\usepackage{booktabs}
\usepackage{algorithm}
\usepackage{algorithmic}
\usepackage{listings}
\newtheorem{example}{Example}

\usepackage{color}
\newcommand{\cred}{\color{red}}
\newcommand{\cblu}{\color{blue}}

\long\def\BOC#1\EOC{\message{(Commented text )}}
\long\def\BOCC#1\EOCC{\message{(Commented text )}}
\long\def\BOCCC#1\EOCCC{\message{(Commented text )}}
\long\def\optional#1{\empty}

\long\def\NBB#1{}

\title{Neuro-Symbolic Reasoning with Large Language Models and Answer Set Programming: A Case Study on Logic Puzzles}

\title{Leveraging Large Language Models to Generate Answer Set Programs}

\author{%
Adam Ishay$^1$\and
Zhun Yang$^1$\and
Joohyung Lee$^{1,2}$\\ 
\affiliations
$^1$Arizona State University\\
$^2$Samsung Research\\
\emails
\{aishay, zyang90, joolee\}@asu.edu
}

\begin{document}
\maketitle

\begin{abstract}
Large language models (LLMs), such as GPT-3 and GPT-4, have demonstrated exceptional performance in various natural language processing tasks and have shown the ability to solve certain reasoning problems. However, their reasoning capabilities are limited and relatively shallow, despite the application of various prompting techniques. In contrast, formal logic is adept at handling complex reasoning, but translating natural language descriptions into formal logic is a challenging task that non-experts struggle with. This paper proposes a neuro-symbolic method that combines the strengths of large language models and answer set programming. Specifically, we employ an LLM to transform natural language descriptions of logic puzzles into answer set programs. We carefully design prompts for an LLM to convert natural language descriptions into answer set programs in a step by step manner. Surprisingly, with just a few in-context learning examples, LLMs can generate reasonably complex answer set programs. The majority of errors made are relatively simple and can be easily corrected by humans, thus enabling LLMs to effectively assist in the creation of answer set programs.
\end{abstract}

\section{Introduction} 

Transformer-based large language models (LLMs) have recently shown remarkable success in many downstream tasks, demonstrating their general reasoning capability across diverse problems.  
However, while LLMs excel in generating System 1 thinking, they struggle with System 2 thinking, resulting in output that is often inconsistent and incoherent \cite{nye21improving}. This is because LLMs are basically trained to predict subsequent words in a sequence and do not appear to have a deep understanding of concepts such as cause and effect, logic, and probability, which are essential for reasoning.

To address the issue, \citeauthor{nye21improving} \shortcite{nye21improving} propose a dual-system model that combines the strengths of LLMs and symbolic logic to achieve improved performance on reasoning tasks. 
They leverage an LLM to generate a System 1 proposal and employ symbolic computation to filter these proposals for consistency and soundness. 


We are interested in situations where problems are described in natural language and solving them requires deep reasoning. A system needs to take into account linguistic variability and be able to perform symbolic reasoning. We take logic puzzles as the testbed as they are well-suited for this purpose. 

We first note that GPT-3 \cite{brown20language} and GPT-4\footnote{Throughout the paper, we use GPT-3 to refer to the ``text-davinci-003'' model and GPT-4 to refer to the ``gpt-4-0314'' (released March, 2023) model in the OpenAI API.} by themselves struggle with solving logic puzzles, despite various prompts we tried. 
On the other hand, we find that they can convert the natural language descriptions of the puzzles into declarative answer set programming languages \cite{lif08,bre11} surprisingly well. 
Even the errors these LLMs make are mostly simple for humans to correct. We hope that our finding will ease the efforts of writing answer set programs and expand the application of answer set programming to a broader audience.

The remainder of this paper is organized as follows. Section~\ref{sec:prelim} offers a brief overview of related work on automated solving of logic puzzles. Sections~\ref{sec:method} and \ref{sec:optional} delve into the proposed approach in detail.
Section~\ref{sec:experiments} presents experimental results and performance evaluations of the approach. Section~\ref{sec:sudoku} shows more examples demonstrating the generalizability of our method. 

The code is available at \url{https://github.com/azreasoners/gpt-asp-rules}.

\section{Preliminaries} \label{sec:prelim}

\subsection{Large Language Models (LLMs)} \label{ssec:llm}

LLMs have significantly improved natural language processing, achieving strong performance on a variety of tasks using few-shot learning \cite{brown20language}. However, LLMs remain weak at tasks that involve complex reasoning \cite{creswell22selection,valmeekam22large}, and scaling model size alone is not enough to achieve good performance \cite{rae21scaling}. It has been shown that various prompting methods improve accuracy on reasoning tasks \cite{wei22chain,zhou22least,creswell22selection}. \citeauthor{nye21improving} \shortcite{nye21improving} present a dual-system model which uses an LLM as a semantic parser and couples it with a custom symbolic module to achieve performance gains on reasoning tasks. This framework combines the strengths of LLMs for parsing complex natural language and symbolic logic for handling complex reasoning. However, the authors had to use hand-engineered set of constraints for the latter part. To our knowledge, our work is the first to use LLMs to generate logic rules to solve complex reasoning tasks.

\subsection{Automated Logic Puzzle Solving} \label{ssec:logic-puzzle}

Works focused on solving logic puzzles typically involve a mapping from natural language to logic formalism. This process often includes problem simplification techniques, such as tailoring the puzzle to a specific domain, restricting natural language input to a certain form, or assuming additional inputs like enumerated types.
\citeauthor{lev04solving} \shortcite{lev04solving} employ a specialized automated multi-stage parsing process to convert natural language text into an intermediate form called Semantic Logic, which is then converted into First Order Logic to finally evaluate on law school admissions tests (LSAT) and the Graduate Records Examination (GRE). \citeauthor{shapiro11jobs} \shortcite{shapiro11jobs} manually encodes the ``Jobs Puzzle'' in a few different logical formalisms and compare them. 
Puzzler \cite{milicevic12puzzler} uses a general link parser to translate puzzles into to the Alloy language for solving, primarily through an automated process, albeit with assumed types. LogicSolver \cite{nordstromlogicsolver} follows a similar approach to Puzzler but replaces Alloy with a custom solver and conducts a more comprehensive evaluation.

Several works utilize translations into the language of answer set programming (ASP) \cite{lif08,bre11}.  \citeauthor{schwitter13jobs} \shortcite{schwitter13jobs} addresses the ``Jobs Puzzle" by representing the problem using controlled natural language \cite{schwitter10controlled}, which can be further turned into ASP. 
\citeauthor{baral12solving} \shortcite{baral12solving} employ a $\lambda$-calculus-based approach and trains a model that converts a manually simplified version of natural language clues into ASP rules for solving Zebra puzzle-type logic puzzles. \citeauthor{mitra15learning} \shortcite{mitra15learning} train a maximum entropy-based model to extract relations for each clue, which are then converted into a common ASP rule format, where a stable model corresponds to the puzzle solution. LGPSolver \cite{jabrayilzade20lgpsolver} uses DistilBERT, a transformer-based model, as a classifier that can distinguish between representative rule types. With the clue classification, the authors use a hand-crafted clue to Prolog translation (as opposed to ASP) and compute the solution. The works mentioned involve some combination of manual processing and/or brittle problem-specific translations. Our work distinguishes itself by being both fully automated and featuring a general pipeline, leveraging the extensive translation capacity available from LLMs.


\subsection{Generate-Define-Test with ASP}  \label{ssec:asp}

ASP programs are typically written following the Generate-Define-Test structure, which generates potential solutions (\textit{Generate}) and eliminates invalid ones based on certain constraints (\textit{Test}). The \textit{Generate} portion usually includes choice rules, while the \textit{Test} portion consists of a set of constraints that prune out invalid solutions. An additional part of the program, the \textit{Define} portion, includes necessary auxiliary predicates that are used in the \textit{Test} portion.

\section{Method}\label{sec:method}

\begin{figure} [ht!]
\centering
\includegraphics[width=0.85\columnwidth]{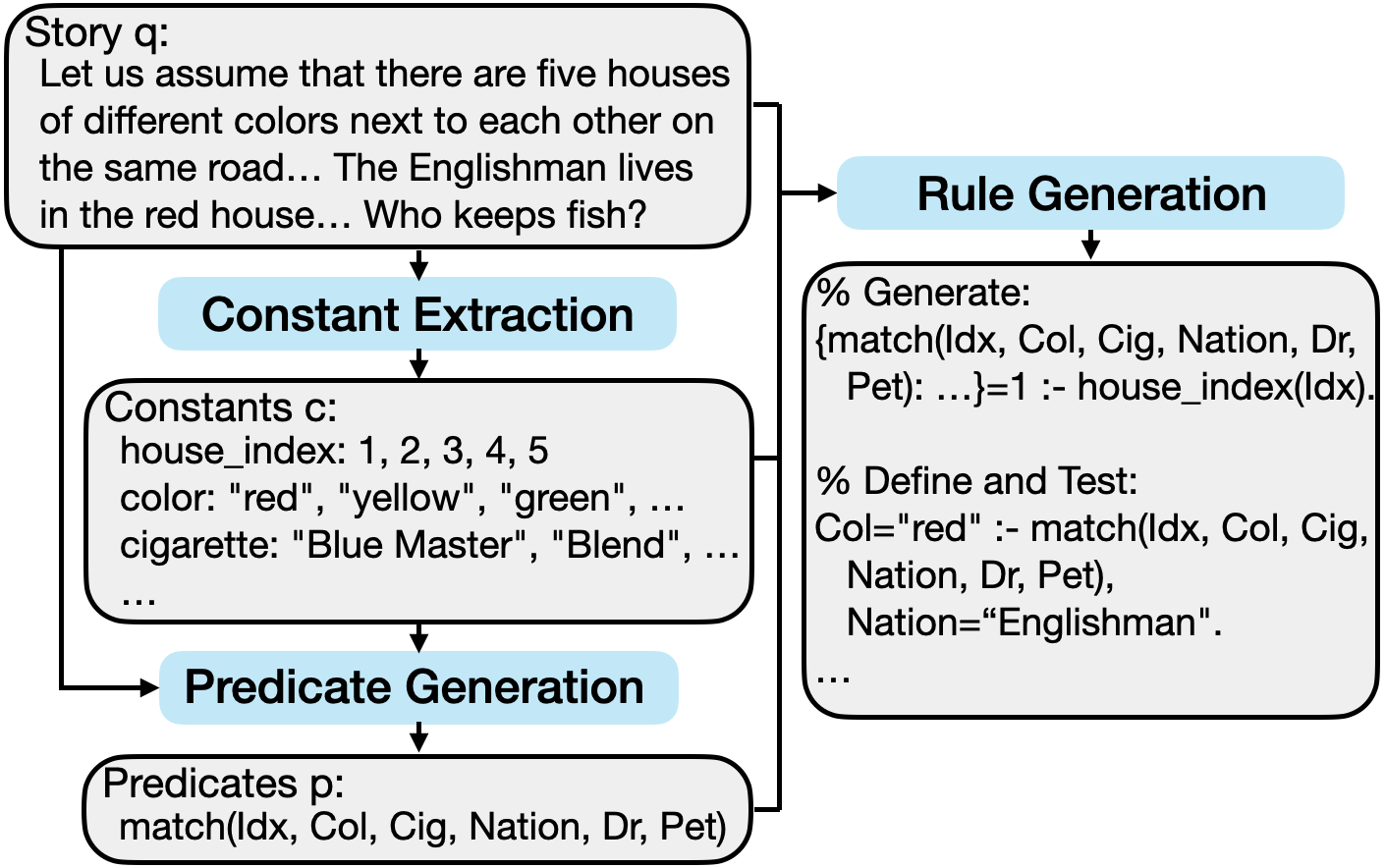}
 \caption{Flow of Generating Answer Set Programs from Logic Puzzle in English} 
 \label{fig:structure}
 \vspace{-4mm}
\end{figure}

In order to find a solution to a logic puzzle, we utilize GPT-3 to convert the puzzle into an answer set program so that the stable model (a.k.a answer set) encodes the solution.\footnote{
Though this section mostly mentions GPT-3, GPT-4 can be used instead.}  Although GPT-3 exhibits strong capabilities, we discovered that it cannot generate a correct answer set program without being guided by carefully engineered prompts. These prompts instructs GPT-3 to reliably extract constants and generate accurate predicates and rules. In this paper, we detail our prompt engineering efforts.

Figure~\ref{fig:structure} illustrates the structure of our pipeline, which utilizes GPT-3 step by step to generate an ASP program. Similar to how a human would approach the task, our pipeline first extracts the relevant objects and their categories. Then, it generates a predicate that describes the relations among the objects from different categories. Using the generated information, the pipeline further constructs an ASP program in the style of Generate-Define-Test.


Let $\mathcal{F}_{c}$ and $\mathcal{F}_{p}$ denote the {\em Constant Extraction} and {\em Predicate Generation} steps in Figure~\ref{fig:structure}. Let $\mathcal{F}_{r1}$ and $\mathcal{F}_{r2}$ represent the two parts of the {\em Rule Generation} step, i.e., the {\em Generate} part and the {\em Define}\&{\em Test} part, respectively.
Our pipeline can be modeled by the following equations that map a puzzle story $q$ to an ASP program $\Pi=\Pi_{generate}\cup \Pi_{define\_test}$.
\begin{align*}
c =& \mathcal{F}_{c}(q) & p =& \mathcal{F}_{p}(q, c) \\
\Pi_{generate} =& \mathcal{F}_{r1}(c, p) & \Pi_{define\_test} =& \mathcal{F}_{r2}(q, c, p).
\end{align*}
Here, $c$ and $p$ denote extracted objects and generated predicates. 
Each step $\mathcal{F}_*$ is realized by GPT-3 with 2-shot prompting, i.e., only 2 examples in each prompt. 

\subsection{Constant Extraction} \label{ssec:constants}


The first step in the pipeline is to extract constants or entities from the given story along with their corresponding categories. To accomplish this, we invoke GPT-3 using {\bf Prompt C}, which consists of three parts: instruction, examples, and a query.


\medskip\noindent
{\bf Prompt C}:
\begin{lstlisting}[escapechar=!]
Given a problem, extract all different constants and their categories in the form "category: constant_1; constant_2; ...; constant_n". Here, the format of each constant is turned into either an integer or a string surrounded by double quotes, e.g.,"some name".

Problem 1:
Consider N-Queens Puzzle on a chessboard of size 8x8. The goal is to assign 8 queens on the chessboard so that no two queens can share the same row, column, or diagonal.

Constants:
index_of_row: 1; 2; 3; 4; 5; 6; 7; 8.
index_of_column: 1; 2; 3; 4; 5; 6; 7; 8.

Problem 2:
"Against the Grain" offers hand-made wooden furniture at reasonable prices. Each item is made by an in-house employee. Using only the clues that follow, match each item to the employee who crafted it, and determine its price and the type of wood used to make it. Remember, as with all grid-based logic puzzles, no option in any category will ever be used more than once.
1. Bonita's piece costs $325.
2. The item made of poplar costs more than Yvette's piece.
3. Tabitha's item costs 50 dollars less than the piece made of sandalwood.
4. The $275 item is either the piece made of ash or Yvette's item.

Constants:
employee: "Bonita"; "Yvette"; "Tabitha".
price: 225; 275; 325.
wood_type: "ash"; "poplar"; "sandalwood".

Problem 3:
!{\cblu <story>}!

Constants:
\end{lstlisting}

Line 1 provides a general instruction for the task of extracting objects and directing GPT-3 to generate them in the form of ``category: constant$_1$; \dots; constant$_n$." Then, two examples follow: Lines 6-8 for Problem 1 specified in Lines 3-4, and Lines 17-20 for Problem 2 specified in Lines 10-15. By replacing Line 23 ({\cblu $\langle$story$\rangle$}) with a new example story and invoking GPT-3 with the above prompt, a new list of categories and constants for that story is generated, as with the previous two examples.

The above two examples are chosen to cover two cases of object extraction. For the N-Queens problem, the constants $1,\dots, 8$ are not described in the Problem 1 statement (Line 4) but can be inferred. For the second puzzle, however, all constants in Lines 18-20 are mentioned in the example story provided in Lines 11-15.



The second puzzle is also intentionally selected to give an example for GPT-3 so that certain constants (e.g., {\tt \$225}) can be turned into valid integers (e.g., {\tt 225}) so that arithmetic can be applied correctly later when generating rules later on, while others should be surrounded by double quotes.
We experimented with various prompts to instruct GPT-3 to generate all non-numeric constants in lowercase and replace special characters with underscores. However, GPT-3 was unable to strictly adhere to these instructions and consequently made more errors.

\BOCC
e find the most success with the following strategy (lines 4-6): require that every object be either an integer or a string surrounded by double quotes; the former (e.g., {\tt 225}) allows for convenient built-in arithmetic computations in {\sc clingo}, and the latter (e.g., {\tt "Sue Simpson"} and {\tt "8:30 AM"}) allows for more free-form content for a object.

{\cblu One way is to supply GPT-3 with a few prompt senteces to turn those objects into so

add a few sentences in the prompt, guiding the LLM to generate all letters in lowercase and replace special characters with underscores.}
However, such translation introduces more errors and doesn't guarantee the validity of the translated object, e.g., {\tt 8\_30\_am}. 

{\cblu We find the most success with the following strategy (lines 4-6): require that every object be either an integer or a string surrounded by double quotes;} the former (e.g., {\tt 225}) allows for convenient built-in arithmetic computations in {\sc clingo}, and the latter (e.g., {\tt "Sue Simpson"} and {\tt "8:30 AM"}) allows for more free-form content for a object.

\medskip\noindent
{Prompt C - instruction}:
\begin{lstlisting}
Given a problem, extract all different objects and
    their categories in the form "category:
    object_1; object_2; ...; object_n". 
    Here, the format of each object is turned into
    either an integer or a string surrounded by
    double quotes, e.g., "some name". 
\end{lstlisting}

\subsubsection{Examples Prompt}
To help GPT-3 to better learn the mapping from a question to the objects we expect to generate, the examples part consists of two examples: N-Queens problem and a logic puzzle from Puzzle Baron~\footnote{\url{https://logic.puzzlebaron.com/}}. 

\medskip\noindent
{Prompt C - examples}:
\begin{lstlisting}[escapechar=^]
Problem 1:
^{\cblu \bf $\langle$question: N-Queens$\rangle$ }^

^{\cblu \bf $\langle$objects: N-Queens$\rangle$ }^

Problem 2:
^{\cblu \bf $\langle$question: logic puzzle$\rangle$ }^

^{\cblu \bf $\langle$objects: logic puzzle$\rangle$ }^
\end{lstlisting}

The prompt above is the examples part of {\bf Prompt C}. Since the same examples are repeatedly used in different prompts, we use $\langle$X$\rangle$ as a shorthand for some specific text identified by the name X as listed below.

\smallskip\noindent
{$\langle$question: N-Queens$\rangle$}:
\begin{lstlisting}
Consider N-Queens Puzzle on a chessboard of size 8x8. The goal is to assign 8 queens on the chessboard so that no two queens can share the same row, column, or diagonal.
\end{lstlisting}

\noindent
{$\langle$objects: N-Queens$\rangle$}:
\begin{lstlisting}
objects:
index_of_row: 1; 2; 3; 4; 5; 6; 7; 8.
index_of_column: 1; 2; 3; 4; 5; 6; 7; 8.
\end{lstlisting}

\noindent
{$\langle$question: logic puzzle$\rangle$}:
\begin{lstlisting}
"Against the Grain" offers hand-made wooden furniture at reasonable prices. Each item is made by an in-house employee. Using only the clues that follow, match each item to the employee who crafted it, and determine its price and the type of wood used to make it. Remember, as with all grid-based logic puzzles, no option in any category will ever be used more than once.
1. Bonita's piece costs $325.
2. The item made of poplar costs more than Yvette's piece.
3. Tabitha's item costs 50 dollars less than the piece made of sandalwood.
4. The $275 item is either the piece made of ash or Yvette's item.
\end{lstlisting}

\noindent
{$\langle$constants: logic puzzle$\rangle$}:
\begin{lstlisting}
Constants:
employee: "Bonita"; "Yvette"; "Tabitha".
price: 225; 275; 325.
wood_type: "ash"; "poplar"; "sandalwood".
\end{lstlisting}

The above two examples are selected to cover two cases of constant extraction. For the N-Queens problem, the constants in $\langle$constants: N-Queens$\rangle$ are not described in the question $\langle$question: N-Queens$\rangle$ but need to be inferred. For the logic puzzle, however, all constants in $\langle$constants: logic puzzle$\rangle$ can be found in $\langle$question: logic puzzle$\rangle$. 

The logic puzzle is also selected intentionally to give an example for GPT-3 that some constants (e.g., {\tt \$225}) need to be turned into a valid integer (e.g., {\tt 225}) and others should be surrounded by double quotes no matter they were valid (e.g., {\tt ash}) or not (e.g., {\tt Bonita}).

\subsubsection{Query Prompt}
The query part below simply follows the same structure as the two examples where $\langle$question: query$\rangle$ is replaced with the queried question $q$. The content of constants $c$ is not provided but is to be generated by GPT-3.

\medskip\noindent
{Prompt C - query}:
\begin{lstlisting}[escapechar=^]
Problem 3:
^{\cblu \bf $\langle$question: query$\rangle$ }^

Constants:
\end{lstlisting}

\EOCC 

\subsection{Predicate Generation} \label{ssec:predicate}

The next step in the pipeline is to generate predicates $p$ that describe the relations among the extracted constants. We use GPT-3 on the {\bf Prompt P} below. 

\medskip
\noindent
{\bf Prompt P}:
\begin{lstlisting}[escapechar=!]
Given a problem and some categorized constants of the form "category: constant_1; constant_2; ...; constant_n", generate the minimum number of predicates to define the relations among the categories of constants. Each generated predicate is of the form "predicate(X1, X2, ..., Xn)" where X1, X2, ..., Xn are different variables and each variable X belongs to one of the categories. For each category, there must exist at least one variable of some predicate that belongs to this category.

Problem 1: 
!{\cblu (Lines 4-8 from Prompt C: Omitted)}!

Predicates:
% The categories in Constants include index_of_row and index_of_column. We use different variables Ir and Ic to represent index_of_row and index_of_column.
% We assign a queen at row Ir and column Ic, where Ir belongs to index_of_row and Ic belongs to index_of_column.
assign(Ir, Ic)

Problem 2: 
!{\cblu (Lines 11-20 from Prompt C: Omitted)}!

Predicates:
% The categories in Constants include employee, price, and wood_type. We use different variables E, P, and W to represent employee, price, and wood_type.
% We match an employee E with price P and wood type W, where E belongs to employee, P belongs to price, and W belongs to wood_type.
match(E, P, W)

Problem 3:
!{\cblu <story>}!

Constants:
!{\cblu <constants> }!
    
Predicates:
\end{lstlisting}


Line 1 is a general instruction describing the task of predicate generation, and that the generated predicates should follow the form of ``predicate(X$_1$, \dots, X$_n$)'' where each X$_i$ is a distinct variable that represents a category of constants. 

Again, the two examples follow. 
%
{
Lines 3--4 are a copy of the first example in Lines 3--8 of {\bf Prompt C} (where we omit Lines 4--8 from {\bf Prompt C} to reduce the space). Lines 6--9 continue the first example, where it now generates the predicates with variables as arguments following the instruction. It also contains two comments (starting with symbol {\tt \%}).
}
The first comment in Line 7 recalls the categories of constants and assigns a different variable to each category. 
{The second comment in Line 8 gives the English reading of the predicate and variables, and emphasizes the link between each variable and a category of constants.}
Similarly, 
{Lines 11--17} present the second example. 

Next, the story and constants are given for the third problem and GPT-3 is prompted to generate the predicate for that example, given the general instruction and the preceding two examples.

Given the extracted constants $c$ and generated predicates $p$, the next step in the pipeline is to generate ASP rules $\Pi$, consisting of the \textit{Generate} part and the \textit{Define\&Test} part.

\subsection{Rule Generation: Generate} \label{ssec:generate}


The \textit{Generate} part of an ASP program defines all possible mappings of constants from different categories. This is done by choice rules.
In this step, an ASP program $\Pi_{generate}$ is obtained by calling GPT-3 with {\bf Prompt R1}. 

\medskip\noindent
{\bf Prompt R1}:
\begin{lstlisting}[escapechar=!]
Given some categorized constants in the form "category: constant_1; constant_2; ...; constant_n" and some predicates about the relation among different categories of constants, write ASP (Answer Set Programming) rules to generate the search space of possible relations.

Constants:
employee: "Bonita"; "Yvette"; "Tabitha".
price: 225; 275; 325.
wood_type: "ash"; "poplar"; "sandalwood".

Predicates:
% The categories include employee, price, and wood_type. We use different variables E, P, and W to represent employee, price, and wood_type.
% We match an employee E with price P and wood type W, where E belongs to employee, P belongs to price, and W belongs to wood_type.
match(E, P, W)

ASP Rules:
% Define the constants in each category.
employee("Bonita"; "Yvette"; "Tabitha").
price(225; 275; 325).
wood_type("ash"; "poplar"; "sandalwood").

% For each employee E, it matches with exactly 1 price P and 1 wood type W.
{match(E, P, W): price(P), wood_type(W)}=1 :- employee(E).

Constants:
!{\cblu <constants>}!

Predicates:
!{\cblu <predicates>}!

ASP rules:
\end{lstlisting}

In the above prompt, {\cblu $\langle$constants$\rangle$} and  {\cblu $\langle$predicates$\rangle$} are to be replaced for a new example. GPT-3 generates facts and choice rules following the last line of the prompt.



\BOC
\medskip\noindent
{Prompt R1 - instruction}:
\begin{lstlisting}
Given some categorized constants in the form 
    "category: constant_1; constant_2; ...; 
    constant_n" and some predicates about the
    relation among different categories of constants,
    write ASP (Answer Set Programming) rules to
    generate the search space of possible relations.
\end{lstlisting}
\EOC

\BOC
\subsubsection{Examples Prompt}
To specify the search space of the stable models, we treat each predicate as a function, whose domain and codomain are determined by the categorized constants. 
A domain (and similarly for codomain), e.g., $\{a, b, c\}$, can be defined by a {\em fact}
\[
domain(a; b; c)
\]
which says that ``$domain(X)$ is true for $X\in\{a,b,c\}$.''
A function $f$ from its domain to codomain can be defined by a {\em choice rule} 
\[
\{f(X,Y): codomain(Y)\}=1 \leftarrow domain(X)
\]
which says ``for every $X$ in the domain, there is exactly one $Y$ in the codomain such that $f(X,Y)$ is true.''~\footnote{Although not needed for our main experiments, this choice rule can be generalized by removing ``$=1$'' or replacing it with ``$=k$'', which works for the cases when a predicate cannot be treated as a function (e.g., in the Jobs Puzzle in Appendix~\ref{apx:job_puzzle}).}
\EOC 

The task in this step is to write facts and choice rules based on the generated constants and predicates. 
Since this step doesn't require the details of the story, we omit the story from the prompt to avoid unnecessary noisy information being included in the prompt. Each example only consists of constants, predicates, and ASP rules to be generated, i.e., facts and choice rules. 

\BOC
\medskip\noindent
{Prompt R1 - examples}:
\begin{lstlisting}[escapechar=^]
^{\cblu \bf $\langle$constants: N-Queens$\rangle$ }^

^{\cblu \bf $\langle$predicates: N-Queens$\rangle$ }^

^{\cblu \bf $\langle$rules\_R1: N-Queens$\rangle$ }^

^{\cblu \bf $\langle$constants: logic puzzle$\rangle$ }^

^{\cblu \bf $\langle$predicates: logic puzzle$\rangle$ }^

^{\cblu \bf $\langle$rules\_R1: logic puzzle$\rangle$ }^
\end{lstlisting}
\EOC


Similar to the previous prompts, Line 1 is a general instruction, Lines 3--20 provide an example, and Lines 22--28 are for the queried example. The example ASP rules in Lines 14--20 contain comments (Lines 14 and 19), which will also be generated for the queried example and help to gather semantic information before generating a rule.

\subsection{Rule Generation: Define and Test} \label{ssec:test}

The \textit{Define\&Test} part of an ASP program contains 
constraints that ``weed out'' the stable models that do not correspond to valid answers. This step takes as input the puzzle story $q$, constants $c$, and predicates $p$: semantically, the ASP rules represent the content in story $q$ while, syntactically, the ASP rules must be formed by the extracted constants $c$ and generated predicates $p$.
The ASP program $\Pi_{define\_test}$ is obtained by calling GPT-3 with {\bf Prompt R2}.

\medskip\noindent{\bf Prompt R2}:
\begin{lstlisting}[escapechar=^]
Consider the constraint in the following form
<C1>; <C2>; ...; <Cm> :- <L1>, <L2>, ..., <Ln>.
which says that if the conjunction "<L1> and <L2> and ... and <Ln>" is true, then the disjunction of comparisons "<C1> or <C2> or ... or <Cm>" must be true.

One can also add a restriction that "exactly k of <C1>, <C2>, ..., <Cm> is true" by using the following form
{<C1>; <C2>; ...; <Cm>}=k :- <L1>, <L2>, ..., <Ln>.

Given a problem, extract all constraints from the clues in the problem using only the provided constants and predicates.

Problem 1:
"Against the Grain" offers hand-made wooden furniture at reasonable prices. Each item is made by an in-house employee. Using only the clues that follow, match each item to the employee who crafted it, and determine its price and the type of wood used to make it. Remember, as with all grid-based logic puzzles, no option in any category will ever be used more than once.
1. Bonita's piece costs $325.
2. The item made of poplar costs more than Yvette's piece.
3. Tabitha's item costs 50 dollars less than the piece made of sandalwood.
4. The $275 item is either the piece made of ash or Yvette's item.

Constants:
employee: "Bonita"; "Yvette"; "Tabitha".
price: 225; 275; 325.
wood_type: "ash"; "poplar"; "sandalwood".

Predicates:
% The categories include employee, price, and wood_type. We use different variables E, P, and W to represent employee, price, and wood_type.
% We match an employee E with price P and wood type W, where E belongs to employee, P belongs to price, and W belongs to wood_type.
match(E, P, W)

Constraints:
% No option in any category will ever be used more than once.
{E1=E2; P1=P2; W1=W2}=0 :- match(E1,P1,W1), match(E2,P2,W2), (E1,P1,W1)!=(E2,P2,W2).

% 1. Bonita's piece costs $325.
P=325 :- match(E,P,W), E="Bonita".

% 2. The item made of poplar costs more than Yvette's piece.
P1>P2 :- match(E1,P1,W1), match(E2,P2,W2), W1="poplar", E2="Yvette".

% 3. Tabitha's item costs 50 dollars less than the piece made of sandalwood.
P1=P2-50 :- match(E1,P1,W1), match(E2,P2,W2), E1="Tabitha", W2="sandalwood".

% 4. The $275 item is either the piece made of ash or Yvette's item.
{W="ash"; E="Yvette"}=1 :- match(E,P,W), P=275.

^{\cblu (Problem 2 omitted)}^

Problem 3:
^{\cblu <story>}^ 

Constants:
^{\cblu <constants>}^

Predicates:
^{\cblu <predicates>}^

Constraints:
\end{lstlisting}
In the above prompt, {\cblu $\langle$story$\rangle$} is a new puzzle, and {\cblu $\langle$constants$\rangle$},  {\cblu $\langle$predicates$\rangle$} are generated by GPT-3 for that story using {\bf Prompt C} and {\bf Prompt P} in Section~\ref{ssec:constants} and \ref{ssec:predicate}.

Lines 1--8 are a general instruction describing the task of $\Pi_{define\_test}$ generation and provides two rule forms for the target ASP rules. 
The first rule form 
\[
C_1; C_2; \dots; C_m ~\leftarrow~ L_1, L_2, \dots, L_n
\]
says that ``$C_1$ or ... or $C_m$ is true if $L_1$ and ... and $L_n$ are true.'' Here, each $L_i$ is a literal and each $C_i$ is a comparison in the input language of {\sc clingo}, e.g., $A>B$, $A=B+3$, etc.
The second rule form
\[
\{C_1; C_2; \dots; C_m\}=k ~\leftarrow~ L_1, L_2, \dots, L_n
\]
additionally restricts that ``exactly $k$ of $\{C_1, \dots, C_m\}$ must be true.''
In principle, the first rule form is enough to represent various constraints. However, since the second rule form is syntactically closer to certain complex sentences related to cardinality, e.g., ``either ... or ...'', ``neither ... nor ...'', or ``no ... is ...'', etc, we found that GPT-3 works much better when we also include the second rule form. 

\section{Optional Enhancements to the Pipeline}  \label{sec:optional}

Section~\ref{sec:method} presented a general pipeline that automatically writes an ASP program for a 
puzzle in natural language using LLM. 
This section explains two optional enhancements that strengthen its robustness.

\subsection{Constant Formatting}


In the Constant Extraction step (Section~\ref{ssec:constants}), 
GPT-3 may extract the names of the objects as they appear in the puzzle story, such as {\tt \$225}, {\tt Sue Simpson}, and {\tt 8:30 AM}, which do not conform to the syntax of the input language of answer set solver {\sc clingo}. 
Also, GPT-3 applies arithmetic computations (e.g., {\tt L1=L2+3}) to constants surrounded by double quotes (e.g., {\tt L2} is constant {\tt "9 inches"}) instead of constants that are integers (e.g., {\tt L2} is constant {\tt 9}).

A rule-based post-processing could be applied to turn them into the right syntax, but alternatively, we employ GPT-3 to generate syntactically correct forms. 
We found that this method requires significantly less efforts and is more general because GPT-3 applies the constant formatting correctly even for unforeseen formats using some “common sense,” which is lacking in the rule-based approach.
We use the following prompt for this. 


The {\em Constant Formatting} step is done by calling GPT-3 with the following prompt, where {\cblu $\langle$constants$\rangle$} at the end of the prompt is replaced by the original (extracted) constants~$c$ obtained by the Constant Extraction step (Section~\ref{ssec:constants}).
The GPT-3 response in this step is the updated constants~$c$, serving as an input to other steps in the pipeline.

\begin{lstlisting}[escapechar=^]
Given categorized constants of the form "category: constant_1; constant_2; ...; constant_n", format the category and constants such that:
each category consists of only lowercase letters and underscores, and
each constant is either an integer or a string surrounded by double quotes, e.g., "United States".

There are two ways below to format constants and we must use the same way for all constants of the same category.
1. Turn all constants of the same category into integers with no space or special character.
2. Add double quotes around all constants of the same category.
Note that the 1st way has a higher priority, meaning that we must turn all constants of the same category into integers whenever possible. For example, twice or second can be turned into 2, September can be turned into 9, September 5th can be turned into 5 if all dates are in September, but 9:30am can only be turned into "9:30am" since no integer can represent 9:30am.

Original constants:
Employees: Bonita; Yvette; Tabitha.
Prices: $225; $275; $325.
Wood types: ash; poplar; sandalwood.

Formatted constants:
employee: "Bonita"; "Yvette"; "Tabitha".
price: 225; 275; 325.
wood_type: "ash"; "poplar"; "sandalwood".

Original constants:
months: January; April; October; December.
times: 8:30AM; 10:30AM; 2:30PM; 3:30PM.
durations: 1 day; 3 days; 11 days; 12 days.

Formatted constants:
month: 1; 4; 10; 12.
time: "8:30AM"; "10:30PM"; "2:30PM"; "3:30PM".
duration: 1; 3; 11; 12.

Original constants:
^{\cblu $\langle$constants$\rangle$ }^

Formatted constants:
\end{lstlisting}

\subsection{Sentence Paraphrasing}

Sometimes sentences may need to be paraphrased before an LLM can correctly generate rules from them. The {\em Sentence Paraphrasing} step provides the opportunity to not only simplify or formalize the sentences from the original question but also add the hidden information assumed to underlie the question. For example, the following sentence
\begin{lstlisting}
Of the person who won the prize in bioengineering and Sue Simpson, one won in 1976 and the other won in 1968.
\end{lstlisting}
is one clue in the example question in Section~\ref{sec:method}. 
The correct translation requires an LLM to turn the above sentence into at least 3 ASP rules, which would be hard for the current LLMs (e.g., GPT-3). Instead, we can ask GPT-3 to first paraphrase such kind of sentence into simpler ones below.
\begin{lstlisting}
The person who won the prize in bioengineering and Sue Simpson are different.
The person who won the prize in bioengineering won in 1976 or won in 1968.
Sue Simpson won in 1976 or won in 1968.
\end{lstlisting}

The Sentence Paraphrasing step is done by calling GPT-3 with the following prompt, where {\cblu $\langle$sentences$\rangle$} at the end of the prompt is replaced by the numbered sentences in the queried puzzle story $q$, and the GPT-3 response in text is used to replace the original sentences in $q$. 
This prompt is dedicated to the logic puzzles from Puzzle Baron and only paraphrases one kind of sentence in the form ``of A and B, one is C and the other is D.''

\begin{lstlisting}[escapechar=^]
Copy a sequence of numbered sentences.

If a sentence is of the form "N. Of A and B, one is C and the other is D", replace it with 3 sentences below.
N.1 A and B are different.
N.2 A is C or D.
N.3 B is C or D.

For every sentence, if it is not of the form "N. Of ... and ...", simply copy it without replacement. An easy way to determine if a sentence is not of the above form is to check if its first word is not of.

In the following example, one sentence is of the above form.
Given:
1. The squad from Grenada ended with 2 silver medals.
2. Of the team from Oman and the team that won 10 silver medals, one finished with 2 gold medals and the other finished with 1 gold medal.
Copy:
1. The squad from Grenada ended with 2 silver medals.
2.1 The team from Oman and the team that won 10 silver medals are different.
2.2 The team from Oman finished with 2 gold medals or finished with 1 gold medal.
2.3 The team that won 10 silver medals finished with 2 gold medals or finished with 1 gold medal.

In the following example, no sentence is of the above form.
Given:
1. Tabitha's item costs 50 dollars less than the piece made of sandalwood.
2. The $275 item is either the piece made of ash or Yvette's item.
Copy:
1. Tabitha's item costs 50 dollars less than the piece made of sandalwood.
2. The $275 item is either the piece made of ash or Yvette's item.

Given:
^{\cblu $\langle$sentences$\rangle$ }^
Copy:
\end{lstlisting}

\section{Experiments} \label{sec:experiments}



We tested the above pipeline on the logic puzzles dataset from \cite{mitra15learning}. Since the constants are provided in the dataset as necessary information to solve each puzzle, we apply Constant Formatting directly on the given constants to generate constants $c$.

The dataset consists of 50 training examples and 100 testing examples. When designing our prompts, we only consult the training examples and not the testing examples. Table~\ref{tab:acc:150puzzles} shows the performance of our approach to zero-shot GPT-3/GPT-4, few-shot GPT-3/GPT-4, and a fully-supervised learning system LOGICIA \cite{mitra15learning}.
\footnote{For GPT-3/GPT-4, to avoid randomness, we use a temperature of 0 (deterministic) and a top P value of 1 (default setting).}
In the few-shot setting, we use the first two examples in the training set as the few-shot examples.
GPT-3 with zero-shot and few-shot settings didn't perform well, while zero-shot GPT-4 could solve 21\% of the test puzzles correctly, which is significantly better than GPT-3's performance. However, this is much lower than our method's 81\%. Interestingly, while the few-shot setting slightly improves over the zero-shot for GPT-3, this is quite different with GPT-4. This is likely because GPT-4 with the zero-shot setting was instructed to solve the puzzles in a step by step manner.
However, for the few-shot setting, the examples only include the problem and solution, which may have discouraged GPT-4 from working through the puzzles in steps.

\begin{table}
\centering
\begin{tabular}{lrr}
\toprule
Method  & train set & test set \\
\midrule
\cite{mitra15learning} & --  & 71\%     \\
Zero-shot GPT-3 & 0\% & 2\%     \\
Few-shot GPT-3 & 4\%  & 3\%     \\
Zero-shot GPT-4 & 12\% & 21\%     \\
Few-shot GPT-4 & 6\%  & 7\%     \\
GPT-3 Generated ASP Rules & 86\%  & 81\%      \\
GPT-4 Generated ASP Rules & {\bf 92}\%  & {\bf 92}\%      \\
\bottomrule
\end{tabular}
\caption{Accuracy on 50 train and 100 test puzzles. GPT-3 refers to the model named``text-davinci-003'' in the OpenAI API, while GPT-4 is the model named ``gpt-4.''}
\label{tab:acc:150puzzles}
\end{table}



\begin{table}
\centering
\begin{tabular}{lcc}
\toprule
 Step & \multicolumn{2}{c}{Count} \\
 & GPT-3 & GPT-4 \\
\midrule
constant formatting & 3 & 1    \\
paraphrasing & 2 & 4  \\
constraint generation (syntax) & 3 & 0\\
constraint generation (semantics) & 13 & 3\\
\bottomrule
\end{tabular}
\caption{Mistakes on 100 test puzzles at different pipeline steps. 
}
\label{tab:mistakes:100puzzles}
\end{table}

Besides the fact that the direct execution of the LLMs results in low performance, it is hard to understand why they fail to solve puzzles; in other words, the results are hard to interpret. 

On the other hand, in our method, although the LLMs can still produce incorrect ASP programs, the errors are much less and it is easy to identify the failures by examining the generated ASP programs. 
Table~\ref{tab:mistakes:100puzzles} summarizes the number of mistakes we encountered on 100 test data at each pipeline step. 
Below we provide a representative example for each type of errors from Table~\ref{tab:mistakes:100puzzles} and highlight the mistakes in {\cred red}. \footnote {We have provided a breakdown of the errors we encountered on Github at \url{https://github.com/azreasoners/gpt-asp-rules/tree/main/error_analysis}.}  The examples are taken from GPT-3, although each type also occurs with GPT-4, except for the syntax error.

\begin{itemize}
\item {\bf Failure to convert certain constants into integers during constant formatting.~~} The only mistake happens when, given constants ``points: 181 points; 184 points; 187 points; 190 points'', GPT-3 generated
\begin{lstlisting}[escapechar=^]
^{\cred points: "181 points"; "184 points";}^
^{\cred "187 points"; "190 points".}^
\end{lstlisting}
while it should generate constants in integers below to be able to apply arithmetic operations.
\begin{lstlisting}
points: 181; 184; 187; 190.
\end{lstlisting}

\item {\bf Addition of wrong clues during paraphrasing.~~} One such error happens when paraphrasing the sentence ``2. The book published in 2011 is either Orlando's project or Jan's project''. GPT-3 generated the following three sentences.


\begin{lstlisting}[escapechar=^]
% 2.1 The book published in 2011 is either Orlando's project or Jan's project.
^{\cred \% 2.2 Orlando's project is the book}^
    ^{\cred published in 2011 or the 200 page}^
    ^{\cred project.}^
^{\cred \% 2.3 Jan's project is the book}^
^{\cred published in 2011 or the 200 page}^
^{\cred project.}^
\end{lstlisting}
The first sentence is simply a copy of the original sentence while the last two should not be there since they cannot be inferred from the original sentence. 

\item {\bf Syntax error during constraint generation.~~}
One typical syntax error is an incorrect variable name in a generated rule. For example, given the sentence ``The study on the Liberty starts 2 months after Eunice's project'', GPT-3 generates the following rule
\begin{lstlisting}[escapechar=^]
M=M1+2 :- match(S,M,Sh), match(S1,M1,Sh1), ^{\cred S="Eunice"}^, Sh="Liberty".
\end{lstlisting}
while the variable {\tt S} in the second line should have been {\tt S1}.

\item {\bf Semantic error during constraint generation.~~}
One typical semantic error is caused by a wrong equation. For example, given the sentence ``the \$35,000 structure is 15 sq ft smaller than the \$29,000 home'', GPT-3 generated
\begin{lstlisting}[escapechar=^]
^{\cred S1=S2+15}^ :- match(C1,S1,P1), match(C2,S2,P2), P1=35000, P2=29000.
\end{lstlisting}
while the equation in the head should have been {\tt S1=S2-15}.
\end{itemize}

While our pipeline doesn't achieve 100\% accuracy on generated ASP programs, most failed puzzles only have one mistake and such a mistake is easy to correct. This means that our pipeline could serve as a good suggestion tool to prepare draft ASP programs for users.
For example, compared to designing all the ASP programs for 50+100 puzzles manually, it would save a significant amount of time to only check the correctness of the automatically generated rules for the programs that don't have a single stable model.

\NBB{Put Examples; give itemized list of the category with some examples. Done.}


\NBB{code is available at [github]. Done.}

\section{More Examples} \label{sec:sudoku}

Previous approaches that automate logic puzzle solving either only predict constants and relations \cite{mitra15learning} or treat rule generation as a classification problem on a small set of rule templates \cite{jabrayilzade20lgpsolver}. 
In comparison, our method is generative, where rules are generated in an open-ended manner under the guidance of a few examples. 

While it's hard to apply the previous methods to other domains without substantial changes, applying our pipeline to new domains requires only minor adjustments on the prompts. To apply our pipeline to other domains, we make a slight adjustment by turning the last sentence in Line 11 of {\bf Prompt R2} into a numbered clue ``0. No option in any category will ever be used more than once.'', since it was specific to grid logic puzzles.

In the following part of this section, we show how our pipeline can be further applied to generate ASP programs for Sudoku and the Jobs Puzzle.

\subsection{Sudoku}\label{ssec:sudoku}

If we describe Sudoku problem with the following story $q$
\begin{lstlisting}
The Sudoku problem is a classic computer science problem that involves filling a 9x9 grid with digits from 1 to 9.
1. The same number cannot appear twice in the same row.
2. The same number cannot appear twice in the same column.
3. The same number cannot appear twice in the 3x3 box. (You can use division / to check if 2 cells are in the same box.)
\end{lstlisting}
our pipeline generates the following ASP program $\Pi$.
\begin{lstlisting}[escapechar=^]
% Define the constants in each category.
index_of_row(1; 2; 3; 4; 5; 6; 7; 8; 9).
index_of_column(1; 2; 3; 4; 5; 6; 7; 8; 9).
number(1; 2; 3; 4; 5; 6; 7; 8; 9).
% For each row Ir and column Ic, it assigns exactly 1 number N.
{assign(Ir, Ic, N): number(N)}=1 :- index_of_row(Ir), index_of_column(Ic).

% 1. The same number cannot appear twice in the same row.
{N1=N2}=0 :- assign(Ir,Ic1,N1), assign(Ir,Ic2,N2), (Ic1,N1)!=(Ic2,N2).

% 2. The same number cannot appear twice in the same column.
{N1=N2}=0 :- assign(Ir1,Ic,N1), assign(Ir2,Ic,N2), (Ir1,N1)!=(Ir2,N2).

% 3. The same number cannot appear twice in the 3x3 box.
{N1=N2}=0 :- assign(Ir1,Ic1,N1), assign(Ir2,Ic2,N2), ^{\cred (Ir1/3,Ic1/3)=(Ir2/3,Ic2/3)}^, (Ir1,Ic1,N1)!=(Ir2,Ic2,N2).
\end{lstlisting}
This ASP program $\Pi$ is almost correct except that the {\cred red} part in Line 16 of $\Pi$ should be
\begin{lstlisting}
((Ir1-1)/3, (Ic1-1)/3) = ((Ir2-1)/3, (Ic2-1)/3)
\end{lstlisting}
since the row and column indices start from 1. This formula seems too difficult for GPT-3 to notice and generate unless some examples are provided . On the other hand, if we slightly adjust Lines 7--8 of {\bf Prompt C} (Section~\ref{ssec:constants}) to make the indices start from 0, then the generated ASP program $\Pi$ becomes correct as Lines 2--3 of $\Pi$ are changed to the following facts.
\begin{lstlisting}
index_of_row(0; 1; 2; 3; 4; 5; 6; 7; 8).
index_of_column(0; 1; 2; 3; 4; 5; 6; 7; 8).
\end{lstlisting}

GPT-4 also fails to generate the last rule correctly, although it makes a different mistake.

\subsection{Jobs Puzzle}
The Jobs Puzzle studied in \cite{schwitter13jobs} asks one to assign 8 different jobs to 4 people while satisfying the given constraints. The full puzzle $q$ is shown below.

\begin{lstlisting}
1. There are four people: Roberta, Thelma, Steve, and Pete.
2. Among them, they hold eight different jobs.
3. Each holds exactly two jobs.
4. The jobs are: chef, guard, nurse, telephone operator, police officer (gender not implied), teacher, actor, and boxer.
5. The job of nurse is held by a male.
6. The husband of the chef is the telephone operator.
7. Roberta is not a boxer.
8. Pete has no education past the ninth grade.
9. Roberta, the chef, and the police officer went golfing together.
Question: Who holds which jobs?
\end{lstlisting}
This puzzle was considered a challenge for logical expressibility and automated reasoning \cite{shapiro11jobs}. 


To apply our method to the Jobs Puzzle, some paraphrasing was needed before the \textit{Define\&Test} part of rule generation. We manually paraphrased the above puzzle to the following
\begin{lstlisting}
There are four people: Roberta, Thelma, Steve, and Pete. Among them, they hold eight different jobs. Each holds exactly two jobs. The jobs are: chef, guard, nurse, telephone operator, police officer (gender not implied), teacher, actor, and boxer. 
5. The job of nurse is held by a male.
6. The husband of the chef is the telephone operator, which means the chef is a female and the telephone operator is a male.
7. Roberta is not a boxer.
8. Pete has no education past the ninth grade. (Three jobs teacher, nurse, and police officer require higher education.)
9. Roberta, the chef, and the police officer went golfing together, meaning they are different people.
10.1 The same job cannot be assigned to different people.
10.2 Roberta is a female.
10.3 Thelma is a female.
10.4 Steve is a male.
10.5 Pete is a male.
10.6 The actor is a male.
Question: Who holds which jobs?
\end{lstlisting}
by turning clues 1--4 as background story, clarifying clues 6, 8, and 9, and adding a few hidden clues numbered 10.X at the end.
%

As for the prompts, we only need to update Line 1 of {\bf Prompt R1} to the following to allow for {\tt \{...\}=k} in a rule.
\begin{lstlisting}
Given a problem as the background information, turn the provided categorized constants and predicates into rules.

Tasks:
1. Write a rule of the form "category(constant_1; constant_2; ...; constant_n)" to represent each row of categorized constants in the form "category: constant_1; constant_2; ...; constant_n".
2. Write a rule of the form "{predicate(X_1, X_2, ..., X_n): category_2(X_2), ..., category_n(X_n)}=k :- category_1(X_1)" for each predicate of the form "predicate(X_1, X_2, ..., X_n)".
\end{lstlisting}

Finally, GPT-3 generates the following ASP program: 
\begin{lstlisting}[escapechar=^]
% Task 1. Define the constants in each category.
person("Roberta"; "Thelma"; "Steve"; "Pete").
job("chef"; "guard"; "nurse"; "telephone operator"; "police officer"; "teacher"; "actor"; "boxer").
gender("male"; "female").
% Task 2. For each person P, there is exactly 2 jobs J and gender G.
{assign(P, J, G): job(J), gender(G)}=2 :- person(P).
% Done tasks 1 and 2.
% 5. The job of nurse is held by a male.
G="male" :- assign(P,J,G), J="nurse".
% 6. The husband of the chef is the telephone operator, which means the chef is a female and the telephone operator is a male.
G1="female" :- assign(P1,J1,G1), assign(P2,J2,G2), J1="chef", J2="telephone operator".
G2="male" :- assign(P1,J1,G1), assign(P2,J2,G2), J1="chef", J2="telephone operator".
% 7. Roberta is not a boxer.
J!="boxer" :- assign(P,J,G), P="Roberta".
% 8. Pete has no education past the ninth grade. (Three jobs teacher, nurse, and police officer require higher education.)
{J="teacher"; J="nurse"; J="police officer"}=0 :- assign(P,J,G), P="Pete".
% 9. Roberta, the chef, and the police officer went golfing together, meaning they are different people.
P1!=P2 :- assign(P1,J1,G1), assign(P2,J2,G2), assign(P3,J3,G3), P1="Roberta", J2="chef", J3="police officer".
P2!=P3 :- assign(P1,J1,G1), assign(P2,J2,G2), assign(P3,J3,G3), P1="Roberta", J2="chef", J3="police officer".
P1!=P3 :- assign(P1,J1,G1), assign(P2,J2,G2), assign(P3,J3,G3), P1="Roberta", J2="chef", J3="police officer".
% 10.1 The same job cannot be assigned to different people.
^{\cred \{P1=P2\}=0 :- assign(P1,J,G1), }^
    ^{\cred assign(P2,J,G2), (P1,G1)!=(P2,G2).}^
% 10.2 Roberta is a female.
G="female" :- assign(P,J,G), P="Roberta".
% 10.3 Thelma is a female.
G="female" :- assign(P,J,G), P="Thelma".
% 10.4 Steve is a male.
G="male" :- assign(P,J,G), P="Steve".
% 10.5 Pete is a male.
G="male" :- assign(P,J,G), P="Pete".
% 10.6 The actor is a male.
G="male" :- assign(P,J,G), J="actor".
\end{lstlisting}
which is almost correct with a single mistake in translating clue 10.1. If we just replace this constraint (in red) with
\begin{lstlisting}
P1=P2 :- assign(P1,J,G1), assign(P2,J,G2).
\end{lstlisting}
the corrected ASP program has exactly one stable model, which is the correct solution to the Jobs Puzzle.

Similarly, GPT-4 also failed to generate a completely correct ASP program. It also couldn't generate a correct rule for constraint 10.1, and furthermore failed to produce the gender category in constant extraction step 
\textbf{Prompt C}), 
 missing ``\texttt{gender: "male"; "female".}''

\section{Conclusion} \label{sec:conclusion}

LLMs are a relatively recent technology that have shown to be disruptive. Despite their wide range of applications, their responses are not always reliable and cannot be trusted.

Automatic rule generation is a difficult problem. However, by using LLMs as a front-end to answer set programming, we can utilize their linguistic abilities to translate natural language descriptions into the declarative language of answer set programs. Unlike previous methods that use algorithmic or machine learning techniques, we find that a pre-trained large language model with a good prompt can generate reasonably accurate answer set programs. We present a pipeline with general steps that systematically build an ASP program in a natural way. This method not only leads to higher accuracy but also makes the results interpretable.

We expect this type of work to expand the application of KR methods that may appear unfamiliar to non-experts. We also anticipate that this pipeline will serve as a suggestion tool to help users prepare valid constants, useful predicates, or draft ASP programs.

\section*{Acknowledgements} 
We are grateful to the anonymous referees for their useful comments.
This work was partially supported by the National Science Foundation under Grant IIS-2006747. 

\bibliographystyle{kr}

\newpage
\appendix

\section{Prompts in the Pipeline} \label{appendix:prompts}
In this section, we list all prompts used in our pipeline.

\medskip\noindent
{Prompt C}:
\begin{lstlisting}
Given a problem, extract all different constants and their categories in the form "category: constant_1; constant_2; ...; constant_n". Here, the format of each constant is turned into either an integer or a string surrounded by double quotes, e.g., "some name".

Problem 1:
Consider N-Queens Puzzle on a chessboard of size 8x8. The goal is to assign 8 queens on the chessboard so that no two queens can share the same row, column, or diagonal.

Constants:
index_of_row: 1; 2; 3; 4; 5; 6; 7; 8.
index_of_column: 1; 2; 3; 4; 5; 6; 7; 8.

Problem 2:
"Against the Grain" offers hand-made wooden furniture at reasonable prices. Each item is made by an in-house employee. Using only the clues that follow, match each item to the employee who crafted it, and determine its price and the type of wood used to make it. Remember, as with all grid-based logic puzzles, no option in any category will ever be used more than once.
1. Bonita's piece costs $325.
2. The item made of poplar costs more than Yvette's piece.
3. Tabitha's item costs 50 dollars less than the piece made of sandalwood.
4. The $275 item is either the piece made of ash or Yvette's item.

Constants:
employee: "Bonita"; "Yvette"; "Tabitha".
price: 225; 275; 325.
wood_type: "ash"; "poplar"; "sandalwood".

Problem 3:
<question: query>

Constants:
\end{lstlisting}

\medskip\noindent
{Prompt P}:
\begin{lstlisting}
Given a problem and some categorized constants of the form "category: constant_1; constant_2; ...; constant_n", generate the minimum number of predicates to define the relations among the categories of constants. Each generated predicate is of the form "predicate(X1, X2, ..., Xn)" where X1, X2, ..., Xn are different variables and each variable X belongs to one of the categories. For each category, there must exist at least one variable of some predicate that belongs to this category.

Problem 1:
Consider N-Queens Puzzle on a chessboard of size 8x8. The goal is to assign 8 queens on the chessboard so that no two queens can share the same row, column, or diagonal.

Constants:
index_of_row: 1; 2; 3; 4; 5; 6; 7; 8.
index_of_column: 1; 2; 3; 4; 5; 6; 7; 8.

Predicates:
% The categories in Constants include index_of_row and index_of_column. We use different variables Ir and Ic to represent index_of_row and index_of_column.
% We assign a queen at row Ir and column Ic, where Ir belongs to index_of_row and Ic belongs to index_of_column.
assign(Ir, Ic)

Problem 2:
"Against the Grain" offers hand-made wooden furniture at reasonable prices. Each item is made by an in-house employee. Using only the clues that follow, match each item to the employee who crafted it, and determine its price and the type of wood used to make it. Remember, as with all grid-based logic puzzles, no option in any category will ever be used more than once.
1. Bonita's piece costs $325.
2. The item made of poplar costs more than Yvette's piece.
3. Tabitha's item costs 50 dollars less than the piece made of sandalwood.
4. The $275 item is either the piece made of ash or Yvette's item.

Constants:
employee: "Bonita"; "Yvette"; "Tabitha".
price: 225; 275; 325.
wood_type: "ash"; "poplar"; "sandalwood".

Predicates:
% The categories in Constants include employee, price, and wood_type. We use different variables E, P, and W to represent employee, price, and wood_type.
% We match an employee E with price P and wood type W, where E belongs to employee, P belongs to price, and W belongs to wood_type.
match(E, P, W)

Problem 3:
<question: query>

<constants: query>

Predicates:
\end{lstlisting}

\medskip\noindent
{Prompt R1}:
\begin{lstlisting}
Given some categorized constants in the form "category: constant_1; constant_2; ...; constant_n" and some predicates about the relation among different categories of constants, write ASP (Answer Set Programming) rules to generate the search space of possible relations.

Constants:
number: 1; 2; 3; 4; 5; 6; 7; 8.

Predicates:
% The categories include number. Note that we must use different variables in each predicate.
% We assign a queen at row N1 and column N2, where N1 belongs to number and N2 belongs to number.
assign(N1, N2)

ASP Rules:
% Define the constants in each category.
number(1; 2; 3; 4; 5; 6; 7; 8).
% For each row N1, there is exactly 1 queen assigned at some column N2.
{assign(N1, N2): number(N2)}=1 :- number(N1).

Constants:
employee: "Bonita"; "Yvette"; "Tabitha".
price: 225; 275; 325.
wood_type: "ash"; "poplar"; "sandalwood".

Predicates:
% The categories include employee, price, and wood_type. We use different variables E, P, and W to represent employee, price, and wood_type.
% We match an employee E with price P and wood type W, where E belongs to employee, P belongs to price, and W belongs to wood_type.
match(E, P, W)

ASP Rules:
% Define the constants in each category.
employee("Bonita"; "Yvette"; "Tabitha").
price(225; 275; 325).
wood_type("ash"; "poplar"; "sandalwood").
% For each employee E, it matches with exactly 1 price P and 1 wood type W.
{match(E, P, W): price(P), wood_type(W)}=1 :- employee(E).

<constants: query>

<predicates: query>

ASP rules:
\end{lstlisting}

\medskip\noindent
{Prompt R2}:
\begin{lstlisting}
Consider the constraint in the following form
<C1>; <C2>; ...; <Cm> :- <L1>, <L2>, ..., <Ln>.
which says that if the conjunction "<L1> and <L2> and ... and <Ln>" is true, then the disjunction of comparisons "<C1> or <C2> or ... or <Cm>" must be true.

One can also add a restriction that "exactly k of <C1>, <C2>, ..., <Cm> is true" by using the following form
{<C1>; <C2>; ...; <Cm>}=k :- <L1>, <L2>, ..., <Ln>.

Given a problem, extract all constraints from the clues in the problem using only the provided constants and predicates.

Problem 1:
"Against the Grain" offers hand-made wooden furniture at reasonable prices. Each item is made by an in-house employee. Using only the clues that follow, match each item to the employee who crafted it, and determine its price and the type of wood used to make it. Remember, as with all grid-based logic puzzles, no option in any category will ever be used more than once.
1. Bonita's piece costs $325.
2. The item made of poplar costs more than Yvette's piece.
3. Tabitha's item costs 50 dollars less than the piece made of sandalwood.
4. The $275 item is either the piece made of ash or Yvette's item.

Constants:
employee: "Bonita"; "Yvette"; "Tabitha".
price: 225; 275; 325.
wood_type: "ash"; "poplar"; "sandalwood".

Predicates:
% The categories include employee, price, and wood_type. We use different variables E, P, and W to represent employee, price, and wood_type.
% We match an employee E with price P and wood type W, where E belongs to employee, P belongs to price, and W belongs to wood_type.
match(E, P, W)

Constraints:
% No option in any category will ever be used more than once.
{E1=E2; P1=P2; W1=W2}=0 :- match(E1,P1,W1), match(E2,P2,W2), (E1,P1,W1)!=(E2,P2,W2).

% 1. Bonita's piece costs $325.
P=325 :- match(E,P,W), E="Bonita".

% 2. The item made of poplar costs more than Yvette's piece.
P1>P2 :- match(E1,P1,W1), match(E2,P2,W2), W1="poplar", E2="Yvette".

% 3. Tabitha's item costs 50 dollars less than the piece made of sandalwood.
P1=P2-50 :- match(E1,P1,W1), match(E2,P2,W2), E1="Tabitha", W2="sandalwood".

% 4. The $275 item is either the piece made of ash or Yvette's item.
{W="ash"; E="Yvette"}=1 :- match(E,P,W), P=275.

Problem 2:
The Winter Olympics have just wrapped up in Norway. Using only the clues that follow, determine the number of gold, silver and bronze medals won by each country. Remember, as with all grid-based logic puzzles, no option in any category will ever be used more than once.
1. The four teams are the team from Bolivia, the team that won 3 gold medals, the team that won 6 silver medals, and the team from Argentina.
2. The team from Oman and the team that won 10 silver medals are different.
3. The team from Oman finished with 2 gold medals or finished with 1 gold medal.
5. The squad that won 3 gold medals, the squad that won 6 silver medals and the squad from Bolivia were all different teams.
6. Neither the team from Argentina nor the squad that won 2 silver medals is the squad that won 4 gold medals.
8. The squad that won 2 gold medals is either the squad that won 6 silver medals or the team from Grenada.

Constants:
country: "Argentina"; "Bolivia"; "Grenada"; "Oman".
silver_medals: 2; 6; 10; 11.
gold_medals: 1; 2; 3; 4.

Predicates:
% The categories include country, silver_medals, and gold_medals. We use different variables C, S, and G to represent country, silver_medals, and gold_medals.
% We assign a country C with silver medals S and gold medals G, where C belongs to country, S belongs to silver_medals, and G belongs to gold_medals.
assign(C, S, G)

Constraints:
% No option in any category will ever be used more than once.
{C1=C2; S1=S2; G1=G2}=0 :- assign(C1,S1,G1), assign(C2,S2,G2), (C1,S1,G1)!=(C2,S2,G2).

% 1. The four teams are the team from Bolivia, the team that won 3 gold medals, the team that won 6 silver medals, and the team from Argentina.
{C="Bolivia"; G=3; S=6; C="Argentina"}=1 :- assign(C,S,G).

% 2. The team from Oman and the team that won 10 silver medals are different.
C1!=C2 :- assign(C1,S1,G1), assign(C2,S2,G2), C1="Oman", S2=10.

% 3. The team from Oman finished with 2 gold medals or finished with 1 gold medal.
{G=2; G=1}=1 :- assign(C,S,G), C="Oman".

% 5. The squad that won 3 gold medals, the squad that won 6 silver medals and the squad from Bolivia were all different teams.
C1!=C2 :- assign(C1,S1,G1), assign(C2,S2,G2), assign(C3,S3,G3), G1=3, S2=6, C3="Bolivia".
C2!=C3 :- assign(C1,S1,G1), assign(C2,S2,G2), assign(C3,S3,G3), G1=3, S2=6, C3="Bolivia".
C1!=C3 :- assign(C1,S1,G1), assign(C2,S2,G2), assign(C3,S3,G3), G1=3, S2=6, C3="Bolivia".

% 6. Neither the team from Argentina nor the squad that won 2 silver medals is the squad that won 4 gold medals.
{C="Argentina"; S=2}=0 :- assign(C,S,G), G=4.
C1!=C2 :- assign(C1,S1,G1), assign(C2,S2,G2), C1="Argentina", S2=2.

% 8. The squad that won 2 gold medals is either the squad that won 6 silver medals or the team from Grenada.
{S=6; C="Grenada"}=1 :- assign(C,S,G), G=2.

Problem 3:
<question: query>

<constants: query>

<predicates: query>

Constraints:
\end{lstlisting}

In {\bf Prompt R2}, since the maximum number of tokens (including the tokens in prompt and response) allowed by GPT-3 is 4096, we deleted 2 clues (numbered 4 and 7) in the question description of the second logic puzzle (lines 44-50) and their corresponding ASP rules (lines 63-85).

\section{A Few Example Answer Set Programs Generated by GPT-3}
\subsection{Puzzle 1 in test set}
Given the question $q$ below,
\begin{lstlisting}
The local foodie club met at Chez Martin last night for their monthly meal. Match each person to their choice of wine and entree, and determine how much each owed at the end of the night. Remember, as with all grid-based logic puzzles, no option in any category will ever be used more than once.
1. The person who had the port paid 1 dollar more than Kurt.
2. Of the person who paid $25 and the person who paid $24, one was Priscilla and the other had the shiraz.
3. Of the person who paid $27 and Priscilla, one had the chianti and the other had the port.
4. Isabel paid $25.
\end{lstlisting}
our pipeline generates the following ASP program
\begin{lstlisting}
% Define the constants in each category.
wine("chianti"; "port"; "riesling"; "shiraz").
price(24; 25; 26; 27).
name("Isabel"; "Kurt"; "Priscilla"; "Robin").
% For each person N, it matches with exactly 1 wine W and 1 price P.
{match(W, P, N): wine(W), price(P)}=1 :- name(N).

% No option in any category will ever be used more than once.
{W1=W2; P1=P2; N1=N2}=0 :- match(W1,P1,N1), match(W2,P2,N2), (W1,P1,N1)!=(W2,P2,N2).

% 1. The person who had the port paid 1 dollar more than Kurt.
P1=P2+1 :- match(W1,P1,N1), match(W2,P2,N2), W1="port", N2="Kurt".

% 2.1 The person who paid $25 and the person who paid $24 are different.
N1!=N2 :- match(W1,P1,N1), match(W2,P2,N2), P1=25, P2=24.

% 2.2 The person who paid $25 was Priscilla or had the shiraz.
{N="Priscilla"; W="shiraz"}=1 :- match(W,P,N), P=25.

% 2.3 The person who paid $24 was Priscilla or had the shiraz.
{N="Priscilla"; W="shiraz"}=1 :- match(W,P,N), P=24.

% 3.1 The person who paid $27 and Priscilla are different.
N1!=N2 :- match(W1,P1,N1), match(W2,P2,N2), P1=27, N2="Priscilla".

% 3.2 The person who paid $27 had the chianti or had the port.
{W="chianti"; W="port"}=1 :- match(W,P,N), P=27.

% 3.3 Priscilla had the chianti or had the port.
{W="chianti"; W="port"}=1 :- match(W,P,N), N="Priscilla".

% 4. Isabel paid $25.
P=25 :- match(W,P,N), N="Isabel".
\end{lstlisting}
which has exactly 1 stable model and this stable model matches with the solution.
\begin{lstlisting}
match("chianti",24,"Priscilla")
match("shiraz",25,"Isabel")
match("riesling",26,"Kurt")
match("port",27,"Robin")
\end{lstlisting}

\subsection{Puzzle 2 in test set}
Given the question $q$ below,
\begin{lstlisting}
A group of friends has decided to try several different weight-loss diets and exercises to see who amongst them can lose the most weight in 3 months. Using only the clues below, match each person to their total number of pounds lost, and determine the diet type they used and their preferred type of exercise. Remember, as with all grid-based logic puzzles, no option in any category will ever be used more than once.
1. Celia used the gluten-free diet.
2. The friend who lost 3 pounds used the low-fat diet.
3. Mandy lost 2 fewer pounds than Raymond.
4. Mandy lost 4 more pounds than Tom.
5. The dieter who used the vegan diet is either Mandy or the friend who lost 3 pounds.
\end{lstlisting}
our pipeline generates the following ASP program
\begin{lstlisting}
% Define the constants in each category.
pounds_lost(3; 5; 7; 9).
diet("dairy-free"; "gluten-free"; "low-fat"; "vegan").
name("Celia"; "Mandy"; "Raymond"; "Tom").
% For each name N, it matches with exactly 1 pounds lost Pl and 1 diet D.
{match(N, Pl, D): pounds_lost(Pl), diet(D)}=1 :- name(N).

% No option in any category will ever be used more than once.
{N1=N2; Pl1=Pl2; D1=D2}=0 :- match(N1,Pl1,D1), match(N2,Pl2,D2), (N1,Pl1,D1)!=(N2,Pl2,D2).

% 1. Celia used the gluten-free diet.
D="gluten-free" :- match(N,Pl,D), N="Celia".

% 2. The friend who lost 3 pounds used the low-fat diet.
Pl=3 :- match(N,Pl,D), D="low-fat".

% 3. Mandy lost 2 fewer pounds than Raymond.
Pl1=Pl2-2 :- match(N1,Pl1,D1), match(N2,Pl2,D2), N1="Mandy", N2="Raymond".

% 4. Mandy lost 4 more pounds than Tom.
Pl1=Pl2+4 :- match(N1,Pl1,D1), match(N2,Pl2,D2), N1="Mandy", N2="Tom".

% 5. The dieter who used the vegan diet is either Mandy or the friend who lost 3 pounds.
{N="Mandy"; Pl=3}=1 :- match(N,Pl,D), D="vegan".
\end{lstlisting}
which has exactly 1 stable model and this stable model matches with the solution.
\begin{lstlisting}
match("Tom",3,"low-fat")
match("Celia",5,"gluten-free")
match("Mandy",7,"vegan")
match("Raymond",9,"dairy-free")
\end{lstlisting}

\subsection{Variants of Sudoku} \label{ssec:-sudoku-variants}

Continuing the process in Section~\ref{sec:sudoku}, 
we can generate the ASP program for variants of Sudoku by adding 1 or 2 more clues to the puzzle description $q$. Below are the newly generated constraints for the added clues in each variant. 

\smallskip
\noindent{Anti-Knight Sudoku}
\begin{lstlisting}
% 4. The same number cannot appear twice in a knight move.
{N1=N2}=0 :- assign(Ir1,Ic1,N1), assign(Ir2,Ic2,N2), |Ir1-Ir2|+|Ic1-Ic2|=3, (Ir1,Ic1,N1)!=(Ir2,Ic2,N2).
\end{lstlisting}

\noindent{Sudoku-X}
\begin{lstlisting}
% 4. The same number cannot appear twice among the cells whose row index is equal to column index.
{N1=N2}=0 :- assign(Ir1,Ic1,N1), assign(Ir2,Ic2,N2), Ir1=Ic1, Ir2=Ic2, (Ir1,Ic1,N1)!=(Ir2,Ic2,N2).

% 5. The same number cannot appear twice among the cells whose row and column indices sum up to 8.
{N1=N2}=0 :- assign(Ir1,Ic1,N1), assign(Ir2,Ic2,N2), Ir1+Ic1=8, Ir2+Ic2=8, (Ir1,Ic1,N1)!=(Ir2,Ic2,N2).
\end{lstlisting}

\noindent{Offset Sudoku}
\begin{lstlisting}
% 4. The same number cannot appear twice among the cells with the same relative position in 3*3 boxes.
{N1=N2}=0 :- assign(Ir1,Ic1,N1), assign(Ir2,Ic2,N2), Ir1\3=Ir2\3, Ic1\3=Ic2\3, (Ir1,Ic1,N1)!=(Ir2,Ic2,N2).
\end{lstlisting}

\section{Additional GPT-4 Analysis} \label{sec:GPT4}

\subsection{Representative Example Errors (for GPT-4)}
\begin{itemize}
    \item {\bf Failure to convert certain constants into integers during constant formatting.~~}
    Given the constants: ``times: 8:00am; 9:00am; 10:00am; 11:00am'', GPT-4 generated
\begin{lstlisting}[escapechar=^]
^{\cred times("8:00am"; "9:00am"; "10:00am"; }^
^{\cred "11:00am").}^
\end{lstlisting}
instead of the correct generation
\begin{lstlisting}
times(8;9;10;11).
\end{lstlisting}
    \item {\bf Addition of wrong clues during paraphrasing.~~}
    Given the clue  ``2. The conductor working on June 12 is either the conductor departing from Buttonwillow or Greg.'', GPT-4 generates the three sentences 
\begin{lstlisting}[escapechar=^]
^{\cred 2.1 The conductor working on June 12 and }^
    ^{\cred Greg are different.}^
2.2 The conductor working on June 12 is either the conductor departing from Buttonwillow or Greg.
^{\cred Greg is either the conductor departing}^
    ^{\cred from Buttonwillow or the conductor}^
    ^{\cred working on June 12}^
\end{lstlisting}
The second sentence is a copy of the original, while 2.1 and 2.3 cannot be inferred and are therefore wrong. 
    \item {\bf Semantic error during constraint generation.~~}
    The sentence ``Vasquez will leave sometime after Macdonald.'' is parsed by GPT-4 into
\begin{lstlisting}[escapechar=^]
^{\cred M1<M2}^ :- schedule(D1,M1,Du1), schedule(D2,M2,Du2), D1="Vasquez", D2="Macdonald".
\end{lstlisting}
which is incorrect, the less than sign should be changed to greater than:
\begin{lstlisting}
M1>M2 :- schedule(D1,M1,Du1), schedule(D2,M2,Du2), D1="Vasquez", D2="Macdonald".
\end{lstlisting}
\end{itemize}

There are no syntax errors encountered with GPT-4.
\subsection{Error Subtypes}
We further break down the paraphrasing error into two types, (p1) a sentence representing an exclusive disjunction is incorrectly translated into additional sentences. For example, ``3. The card with an APR of 11\% is either the card with the \$4,000 credit limit or the one with the \$20,000 credit limit.'', is incorrectly translated into

 \begin{lstlisting}
3.1 The card with an APR of 11% and the card with the $4,000 credit limit are different.
3.2 The card with an APR of 11% is either the card with the $4,000 credit limit or the one with the $20,000 credit limit.
3.3 The card with the $4,000 credit limit is either the card with an APR of 11% or the one with the $20,000 credit limit.
 \end{lstlisting}
and (p2) a sentence representing that four things are different is incorrectly translated into two incorrect sentences. For example, the statement `` 5. The four people are Deep Shadow, the superhero who started in 2007, the hero who started in 2009 and Matt Minkle.'', is incorrectly translated into

\begin{lstlisting}
5.1 Deep Shadow, the superhero who started in 2007, the hero who started in 2009, and Matt Minkle are different.
5.2 Deep Shadow is either the superhero who started in 2007 or the hero who started in 2009 or Matt Minkle.
5.3 The superhero who started in 2007 is either Deep Shadow or the hero who started in 2009 or Matt Minkle.
5.4 The hero who started in 2009 is either Deep Shadow or the superhero who started in 2007 or Matt Minkle.
\end{lstlisting}

Constraint Generation (semantic errors) is further broken into four subtypes. The first (c1) has to do with an incorrect comparison between times. For example, the statement ``Tricia came in a half-hour after Ora.'' is incorrectly translated into

\begin{lstlisting}
% 2. Tricia came in a half-hour after Ora.
T1="7:00am", T2="7:30am" :- order(C1,B1,T1), order(C2,B2,T2), C1="Ora", C2="Tricia".
T1="7:30am", T2="8:00am" :- order(C1,B1,T1), order(C2,B2,T2), C1="Ora", C2="Tricia".
T1="8:00am", T2="8:30am" :- order(C1,B1,T1), order(C2,B2,T2), C1="Ora", C2="Tricia".

\end{lstlisting}
The second (c2) is when an incorrect operator is used (e.g.  ``+'' in place of ``-''). For example, the statement ``\% 1. Vasquez will leave sometime after Macdonald.'' is incorrectly translated into
\begin{lstlisting}
M1<M2 :- schedule(D1,M1,Du1), schedule(D2,M2,Du2), D1="Vasquez", D2="Macdonald".
\end{lstlisting}

the third (c3) is a disjunction in the head of a rule which should not be there. For example, the statement ``\% 3. The 11-year-old bird has a wingspan 8 inches shorter than Charlie.'' is incorrectly translated into

\begin{lstlisting}
A=11, W=W1-8 :- assign(N,W,A), assign(N1,W1,A1), N="Charlie".
\end{lstlisting}

and last (c4) belongs to semantic errors which do not fit into any of the previous types and only occur once.

\begin{table}[h]
\centering
\begin{tabular}{llcc}
\toprule
Error & Subtype &\multicolumn{2}{c}{Count} \\
& & GPT-3 & GPT-4  \\
\midrule
Constant Formatting& & 3 & 1   \\
\hline
Paraphrasing& & 2 & 3  \\
&p1 &  1 & 3  \\
&p2 &  1 & 1\\
\hline
Cons. Gen. (syntax)& & 3 & 0\\
\hline
Cons. Gen. (semantics)& & 13 & 4\\
&c1 &   3 & 2  \\
&c2 &  4 & 1 \\
&c3 &  3 & 0  \\
&c4  & 3 & 0\\
\bottomrule
\end{tabular}
\caption{Errors on 100 test puzzles on GPT-3 and GPT-4, broken down by subtype.}
\label{tab:mistakes_extended}
\end{table}

Table~\ref{tab:mistakes_extended} shows the count for the errors and error subtypes encountered for GPT-3 and GPT-4. We find that unlike GPT-3, GPT-4 does not make any syntax errors, however it increases its rate of paraphrasing errors.

\subsection{Sudoku and Jobs Puzzle}

\textbf{Sudoku}~~ GPT-4 correctly generates all rules except for the the constraints here:

\begin{lstlisting}[escapechar=^]
% 1. The same number cannot appear twice in the same row.
{^{\cred {Ir1=Ir2; Ic1=Ic2;}^ N1=N2}=0 :- assign(Ir1,Ic1,N1), assign(Ir2,Ic2,N2), (Ir1,Ic1,N1)!=(Ir2,Ic2,N2), Ir1=Ir2.
% 2. The same number cannot appear twice in the same column.
{^{\cred Ir1=Ir2; Ic1=Ic2;}^ N1=N2}=0 :- assign(Ir1,Ic1,N1), assign(Ir2,Ic2,N2), (Ir1,Ic1,N1)!=(Ir2,Ic2,N2), Ic1=Ic2.
% 3. The same number cannot appear twice in the 3x3 box.
{^{\cred Ir1=Ir2; Ic1=Ic2;}^ N1=N2}=0 :- assign(Ir1,Ic1,N1), assign(Ir2,Ic2,N2), (Ir1,Ic1,N1)!=(Ir2,Ic2,N2), (Ir1-1)/3=(Ir2-1)/3, (Ic1-1)/3=(Ic2-1)/3.
\end{lstlisting}
The red portion should not be included, otherwise the program runs correctly. Note that GPT-4 generates the correct \texttt{(Ir1-1)}, \texttt{(Ir2-2)}, \texttt{(Ic1-1)}, and \texttt{(Ic2-1)} terms while GPT-3 does not.

\textbf{Jobs Puzzle}~~ In the constant extraction step, GPT-4 fails to generate the gender category. From the problem

\begin{lstlisting}
Problem 3:
1. There are four people: Roberta, Thelma, Steve, and Pete.
2. Among them, they hold eight different jobs.
3. Each holds exactly two jobs.
4. The jobs are: chef, guard, nurse, telephone operator, police officer (gender not
implied), teacher, actor, and boxer.
5. The job of nurse is held by a male.
6. The husband of the chef is the telephone operator.
7. Roberta is not a boxer.
8. Pete has no education past the ninth grade.
9. Roberta, the chef, and the police officer went golfing together.
Question: Who holds which jobs?
\end{lstlisting}
It produces:
\begin{lstlisting}
person: "Roberta"; "Thelma"; "Steve"; "Pete".
job: "chef"; "guard"; "nurse"; "telephone operator"; "police officer"; "teacher"; "actor"; "boxer"
\end{lstlisting}

and is missing ``\texttt{gender: "male"; "female".}''

Also, on prompt \textbf{Prompt R1}, GPT-4 produces the correct output but then continues to produces constraints that it should not:

\begin{lstlisting}[escapechar=^]
% Task 1. Define the constants in each category.
person("Roberta"; "Thelma"; "Steve"; "Pete").
job("chef"; "guard"; "nurse"; "telephone operator"; "police officer"; "teacher"; "actor"; "boxer").
gender("male"; "female").
% Task 2. For each person P, they hold exactly 2 jobs J and have a gender G.
{assign(P, J, G): job(J), gender(G)}=2 :- person(P).
^{\cred \% Additional rules based on the problem}^
    ^{\cred constraints: }^
^{\cred \% 5. The job of nurse is held by a male. }^
^{\cred assign(P, "nurse", "male") :- person(P). }^
^{\cred \% 6. The husband of the chef is the}^
    ^{\cred telephone operator. }^
^{\cred assign(P1, "chef", "female"), assign(P2,}^
    ^{\cred "telephone operator", "male") :-}^
    ^{\cred person(P1), person(P2). }^
^{\cred \% 7. Roberta is not a boxer.}^ 
^{\cred :- assign("Roberta", "boxer", \_). }^
^{\cred \% 8. Pete has no education past the ninth}^
    ^{\cred grade. }^
^{\cred :- assign("Pete", "teacher", \_). }^
^{\cred \% 9. Roberta, the chef, and the police}^
    ^{\cred officer went golfing together. }^
^{\cred assign("Roberta", \_,\_), assign(\_, "chef",\_),}^
    ^{\cred assign(\_, "police officer",\_ ) :-}^
    ^{\cred person().}^
\end{lstlisting}

Since these constraints not supposed to be generated yet, they are not appropriately prompted and hence GPT-4 produces incorrect constraints. Apart from that, GPT-4 produces a similar incorrect rule on rule 10.1 that GPT-3 produces. 

\begin{lstlisting}
% 10.1 The same job cannot be assigned to different people.
{P1=P2}=0 :- assign(P1,J1,G1), assign(P2,J2,G2), J1=J2, P1!=P2.
\end{lstlisting}
which should be 
\begin{lstlisting}
P1=P2 :- assign(P1,J,G1), assign(P2,J,G2).
\end{lstlisting}

\end{document}